\documentclass[letter, 10pt, conference]{IEEEtran}
\usepackage[letterpaper, left=0.75in, right=0.75in, bottom=0.75in, top=1in]{geometry}
\IEEEoverridecommandlockouts

\usepackage{times}

\usepackage[numbers]{natbib}
\usepackage{multicol}
\usepackage[
  bookmarks=true,
  hidelinks
]{hyperref}

\usepackage{amsmath}
\usepackage{amssymb}

\usepackage{cleveref}

\usepackage{graphicx}
\graphicspath{{./figures/}{./eval/}}
\usepackage{grffile}
\usepackage{subcaption}

\usepackage{tikz}
\usetikzlibrary{shapes.geometric, arrows, chains, calc, positioning}
\tikzstyle{sde} = [draw,minimum width=7cm,
minimum height=.7cm,fill=blue!40]
\tikzstyle{ssk} = [draw,minimum width=2cm,
minimum height=.7cm,fill=green!20]
\tikzstyle{sfc} = [draw,minimum width=2cm,
minimum height=.7cm,fill=yellow!20]
\tikzstyle{scm} = [draw,minimum width=2cm,
minimum height=.7cm,fill=orange!20]
\tikzstyle{sen} = [draw,minimum width=7cm,
minimum height=.7cm,fill=red!20]
\tikzstyle{nobox} = [minimum width=.7cm,minimum height=.7cm]

\usepackage{algorithm}
\usepackage{algpseudocode}

\usepackage{booktabs}
\usepackage{tabularx}

\usepackage{todonotes}

\usepackage[nolist, nohyperlinks]{acronym}
\usepackage{listings}
\usepackage{xcolor}
\definecolor{codegreen}{rgb}{0,0.6,0}
\definecolor{codegray}{rgb}{0.5,0.5,0.5}
\definecolor{codepurple}{rgb}{0.58,0,0.82}
\definecolor{backcolour}{rgb}{0.95,0.95,0.92}
\lstdefinestyle{mystyle}{
    backgroundcolor=\color{backcolour},
    commentstyle=\color{codegreen},
    keywordstyle=\color{magenta},
    numberstyle=\tiny\color{codegray},
    stringstyle=\color{codepurple},
    basicstyle=\ttfamily\footnotesize,
    breakatwhitespace=false,
    breaklines=true,
    captionpos=b,
    keepspaces=true,
    numbers=left,
    numbersep=5pt,
    showspaces=false,
    showstringspaces=false,
    showtabs=false,
    tabsize=2
}
\begin{document}

\begin{acronym}
  \acro{AS2FM}{Autonomous Systems to Formal Models}
  \acro{BT}{Behavior Tree}
  \acro{DSL}{Domain Specific Language}
  \acro{HL-SCXML}{High Level SCXML}
  \acro{LL-SCXML}{Low Level SCXML}
  \acro{LTL}{Linear Temporal Logic}
  \acro{MC}{Model Checking}
  \acro{MDP}{Markov Decision Process}
  \acro{ROS 2}{Robot Operating System 2}
  \acro{SCXML}{State Chart XML}
  \acro{SMC}{Statistical Model Checking}
  \acro{STL}{Signal Temporal Logic}
  \acro{TA}{Timed Automaton}
  \acro{VnV}[V\&V]{Verification and Validation}
  \acrodefplural{MDP}{Markov Decision Processes}
  \acrodefplural{TA}{Timed Automata}
\end{acronym}


\title{AS2FM:\ Enabling Statistical Model Checking of ROS 2 Systems for Robust Autonomy}

\author{
    Christian Henkel$^{1,2,\ast}$, Marco Lampacrescia$^{1,3}$, Michaela Klauck$^{1,4}$, Matteo Morelli$^{5}$
    \thanks{
      $^{1} $Bosch Research, Robert Bosch GmbH, Stuttgart, Germany,
      $^{2} $\texttt{hec2le@bosch.com},
      $^{3} $\texttt{lam2rng@bosch.com},
      $^{4} $\texttt{michaela.klauck@de.bosch.com},
      $^{5} $List, CEA, Universite Paris-Saclay, France \texttt{Matteo.MORELLI@cea.fr}.
    }
    \thanks{$^{2,3}$ The first two authors contributed equally to this work.}
    \thanks{$^{\ast}$ Corresponding author.}
    \thanks{The research was funded by the European Union's Horizon Europe Research \& Innovation Program under Grant 101070227 (CONVINCE).}
    \thanks{© 2025 IEEE. Personal use of this material is permitted. Permission from IEEE must be obtained for all other uses, in any current or future media, including reprinting/republishing this material for advertising or promotional purposes, creating new collective works, for resale or redistribution to servers or lists, or reuse of any copyrighted component of this work in other works.}
}

\maketitle

\begin{abstract}
  Designing robotic systems to act autonomously in unforeseen environments is a challenging task.
  This work presents a novel approach to use formal verification, specifically \ac{SMC}, to verify system properties of autonomous robots at design-time.
  We introduce an extension of the SCXML format, designed to model system components including both \ac{ROS 2} and \ac{BT} features.
  Further, we contribute \ac{AS2FM}, a tool to translate the full system model into JANI.
  The use of JANI, a standard format for quantitative model checking, enables verification of system properties with off-the-shelf \ac{SMC} tools.
  We demonstrate the practical usability of AS2FM both in terms of applicability to real-world autonomous robotic control systems, and in terms of verification runtime scaling.
  We provide a case study, where we successfully identify problems in a \ac{ROS 2}-based robotic manipulation use case that is verifiable in less than one second using consumer hardware.
  Additionally, we compare to the state of the art and demonstrate that our method is more comprehensive in system feature support, and that the verification runtime scales linearly with the size of the model, instead of exponentially.
\end{abstract}

\acresetall

\IEEEpeerreviewmaketitle

\def\smcstorm{SMC\_STORM}
\def\jani{JANI}

\newcommand{\code}[1]{\texttt{#1}}

\def\true{\texttt{true}}
\def\false{\texttt{false}}

\section{Introduction}\label{sec:introduction}
Autonomous robotic systems are expected to operate autonomously in unforeseen environments.
Designing such systems is challenging for several reasons.
Firstly, the required software and hardware components have a high inherent complexity.
Secondly, the exact environment in which the robot will operate is not known at design time.
However, it is required for the robot to robustly operate in the environment by using the available hardware and software components.
Here, \emph{robustness} means that the robot operates successfully in a variety of environments and under a variety of conditions.

To handle the complexity of autonomous robotic systems, a skill-based architecture is often used.
Each skill is responsible for a specific task, such as moving the robot, detecting objects, or planning a path~\cite{alboreSkillbasedDesignDependable2023,rovidaSkiROSSkillBasedRobot2017,saukkoriipiProgrammingControlSkillbased2020}.
Responsible for the coordination of the skills is an architectural layer that we call the \emph{deliberation layer}~\cite{ingrandDeliberationAutonomousRobots2017}.
This layer can be implemented using a variety of techniques, such as \acp{BT}~\cite{colledanchiseBehaviorTreesRobotics2018}.
The general architecture and communication of modern robotic systems is often implemented using the \ac{ROS 2} middleware~\cite{quigleyROSOpensourceRobot2009} because it provides a flexible and modular framework for it.

\def\figwidth{.48\columnwidth}
\begin{figure}[t]
    \centering
    \begin{subfigure}[t]{\figwidth}
        \includegraphics[width=\columnwidth]{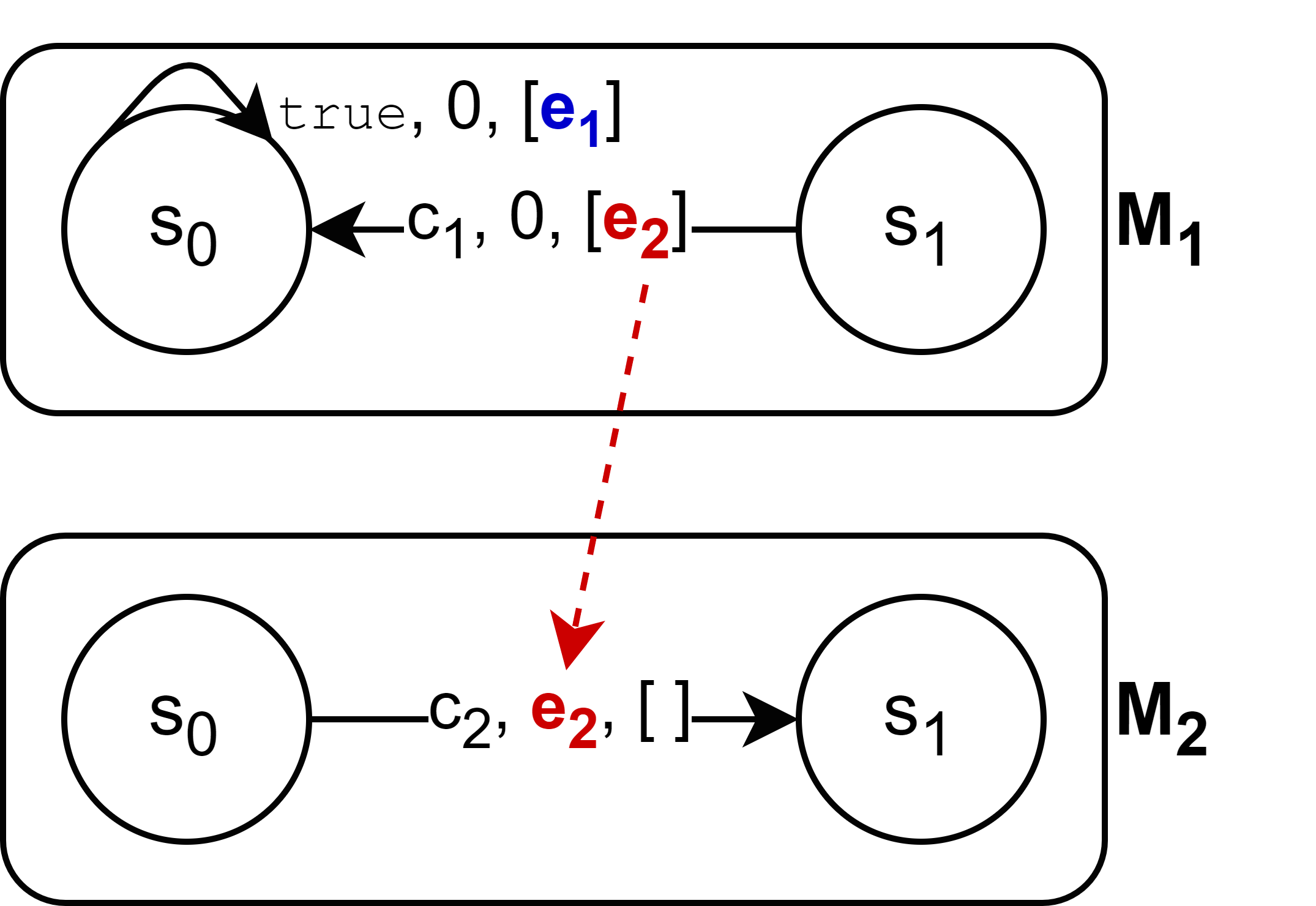}
        \caption{}\label{fig:state-machines}
    \end{subfigure}
    \hfill
    \begin{subfigure}[t]{\figwidth}
        \includegraphics[width=\columnwidth]{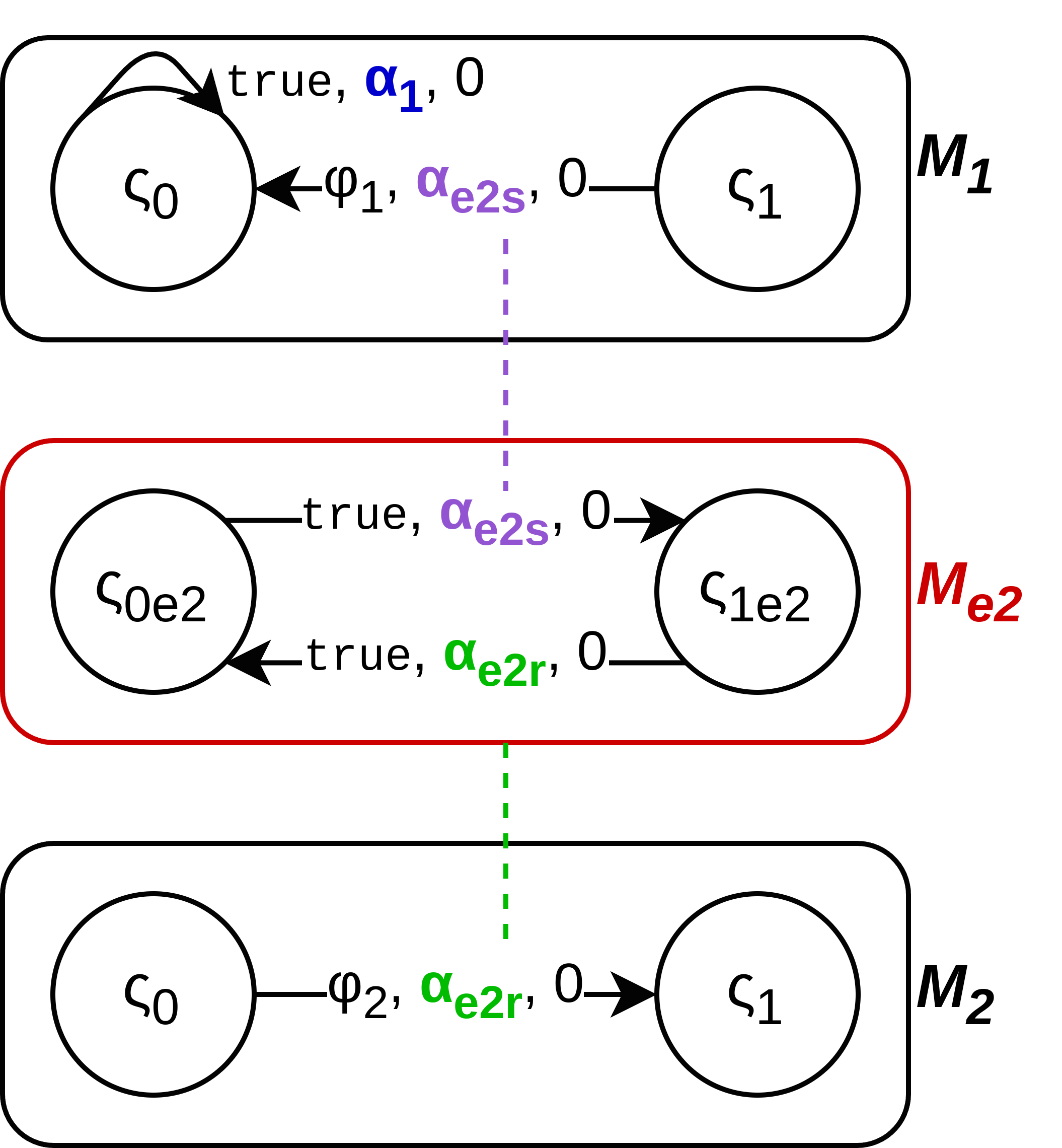}
        \caption{}\label{fig:mdps}
    \end{subfigure}
    \caption{
        Example of two state machines and their translation to a network of \acp{MDP}.\\
        \textbf{(\subref{fig:state-machines})} $M_1$ sends on the transition from $s_1$ to $s_0$ an event $e_2$ highlighted in red that triggers a transition in $M_2$. Note that the event $e_1$ colored in blue is sent without any effect.\\
        \textbf{(\subref{fig:mdps})} Network of \acp{MDP} after translation from the state machines in (\subref{fig:state-machines}). The additional \ac{MDP} $M_{e2}$ in red is used to model the sending and receiving of events. The actions $a_{e2s}$ and $a_{e2r}$ colored in purple and green respectively are synchronized between the \acp{MDP}. Note that the blue action $a_1$ is not synchronized to any other action.
    }
    \vspace*{-0.2cm}
\end{figure}

The use of formal methods, such as model checking, can help to ensure that the robot behaves as expected in all situations. This approach is the core idea of the EU Horizon project CONVINCE~\cite{ERF-2024}.
In model checking, a formal model of the system is used to verify that the system satisfies a given temporal logic property~\cite{baierPrinciplesModelChecking2008}.
However, traditional model checking, also known as full-state space, or exhaustive, model checking, can not handle the complexity of autonomous robotic systems due to the large state space of these systems.
We therefore propose to use \emph{statistical model checking}~\cite{DBLP:series/lncs/LegayLTYSG19,DBLP:journals/sttt/YounesKNP06} to verify properties of autonomous robotic systems.

This paper presents a novel approach to model and verify temporal logic properties of autonomous robotic systems using statistical model checking.
Our tool \ac{AS2FM}, part of the CONVINCE toolbox for enhancing robust robot deliberation~\cite{AAAI-Fall-Symp-2024}, facilitates the translation of complete robot models into a formal model in {\jani}~\cite{DBLP:conf/tacas/BuddeDHHJT17} that can be verified, e.g., using SMC Storm~\cite{lampacresciaVerifyingRoboticSystems2024}.
AS2FM is available open source under Apache 2.0 license\footnote{\url{https://github.com/convince-project/as2fm}}.
To define the complete system, we use \ac{SCXML} to model \ac{ROS 2} nodes, the environment and \ac{BT}-Plugins, and the \ac{BT}-definition of the system \emph{as it is} to model the deliberation layer.

Our work includes in particular:

\begin{itemize}
  \item Syntax and semantics extending \ac{SCXML} to model \ac{ROS 2} communication interfaces, \ac{BT}-specific features, timers, and non-deterministic transitions, and
  \item A translation algorithm from the event-based synchronization of \ac{SCXML} to synchronization of transitions in {\jani} models, implemented together with the rest of the model translation form SCXML to JANI in AS2FM.
\end{itemize}

These contributions help to increase the robustness of autonomous robotic systems by enabling the verification of properties at design-time.
This is to our knowledge the first approach targeting the verification of complete \ac{ROS 2} systems (including communication, timers, and BTs).

The remainder of this paper is structured as follows.
\Cref{sec:related-work} discusses related work.
\Cref{sec:modeling-language} introduces the high-level modeling language used to model robotic systems.
\Cref{sec:model-translation} describes the translation of the high-level model to a formal model.
\Cref{sec:evaluation} presents the model checking evaluation of our approach,
while \Cref{sec:conclusion} summarizes findings and ideas for future work.

\section{Related Work}\label{sec:related-work}

Formal verification is a rigorous mathematical approach used to ensure that the designed system adheres to specified properties and requirements. It proves the correctness of algorithms and systems, particularly in critical applications such as software, hardware, and protocols. One of the key techniques in formal verification is \ac{MC}~\cite{clarke2018handbook}, which systematically explores the state space of a system's model to verify whether it satisfies certain properties, typically expressed in temporal logic~\cite{DBLP:conf/banff/Vardi95}. It provides a powerful tool for validating complex systems against their specifications, ensuring reliability and safety in system design. Traditionally, full-state space (exhaustive) model checking generates traces to explore the entire reachable state space of the model~\cite{baierPrinciplesModelChecking2008}, which is quite resource consuming.
This is the reason for a lack of methods and tools to validate complex robotic systems efficiently \emph{as a whole}, despite the longstanding research interest in \ac{VnV} of autonomous systems~\cite{DBLP:journals/csur/LuckcuckFDDF19}.
A solution to that problem is using \emph{\acf{SMC}}~\cite{DBLP:series/lncs/LegayLTYSG19,DBLP:journals/sttt/YounesKNP06}.
\ac{SMC} evaluates the model by only simulating enough traces, until statistical evidence whether the investigated property holds is achieved.
This is more scalable than full-state space model checking, and it helps finding corner cases faster compared to classic software testing.
However, \ac{SMC} could miss rare edge-cases due to it's probabilistic nature.
Recent work by the authors extend the well-known probabilistic model checker Storm~\cite{DBLP:journals/sttt/HenselJKQV22} with \ac{SMC} for robotic use cases~\cite{lampacresciaVerifyingRoboticSystems2024}.


Storm uses, beside others, \acfp{MDP} as input models, which is a well-known concept in the robotics community.
\acp{MDP} provide a formal framework for modeling decision-making under uncertainty and in dynamic environments~\cite{DBLP:journals/ai/KaelblingLC98,DBLP:conf/ictai/LarocheS99,DBLP:conf/aaai/MatignonJM12,DBLP:journals/ai/IngrandG17}, making them well-suited for representing the complex and dynamic nature of robotic use cases and scenarios.

On the other hand, \acp{TA} are a formalism for modeling and verifying timed systems~\cite{foughaliModelCheckingRealTime2016}.
While \acp{TA} have been used in modeling real-time systems, their application to autonomous robotic decision-making is limited due to the complexity of modeling explicit time constraints~\cite{DBLP:journals/tcs/AlurD94,DBLP:conf/sfm/BehrmannDL04,DBLP:journals/sttt/BehrmannBLP06}.
The tooling for \acp{TA} is not as mature yet as the one for \acp{MDP}~\cite{DBLP:journals/csr/WaezDR13}.
While \acp{TA} are possible in general, their current complexity and the lack of performant tooling make them less practical for modeling complex robotic systems.
Therefore, we propose to use \acp{MDP} for modeling the environment and the system's behavior, as they are more suitable for the complexity of robotic systems.

Previous work has focused on the \ac{VnV} of robotic systems using \ac{MC}~\cite{luckcuckFormalSpecificationVerification2020}, e.g., by verifying entire programs, including scheduling aspects, formalized as timed automata, by compilation to bytecode and translation to UPPAAL for formal analysis~\cite{iversenModelcheckingRealtimeControl2000}.
Other work provides a case study on household robots' behavior definitions that are verified using NuSMV~\cite{dixonFridgeDoorOpenTemporal2014}.
A \ac{DSL} was introduced for specifying the integration of robotic systems to verify the properties of the system using Fiacre and Tina~\cite{foughaliModelCheckingRealTime2016}.
Further, an integrated approach between offline \ac{MC} and online monitoring for robotic systems verifying \ac{STL} properties~\cite{desaiCombiningModelChecking2017} has been published.

Research on an UML-based \ac{DSL} to describe a robotic system and convert it into a TA is presented in~\cite{miyazawaRoboChartModellingVerification2019}.
That work has been continued by introducing a new \ac{DSL} for representing the environment and verifying the resulting \ac{TA} using UPPAAL~\cite{baxterRoboWorldVerificationRobotic2023}.

Some work has been done on formal verification in ROS~1~\cite{hazimTestingVerificationImprovements2016,halderFormalVerificationROSbased2017,carvalhoVerificationSystemwideSafety2020,santosSafetyVerificationRos2021,baxterRoboWorldVerificationRobotic2023}, but we are to our knowledge the first explicitly targeting ROS~2.

There also exists work on verifying \acp{BT}~\cite{serbinowskaFormalizingStatefulBehavior2024}.
The tool BehaVerify focuses on stateful \acp{BT}, which means the \ac{BT} black board is used to keep information in memory.
In that approach the \ac{BT} is converted to an \ac{MDP}, that can be verified using nuXmv.
Though we think this work is the closest to ours, no explicit model of the environment is provided there, since it is described using the black board, and there is no way to describe a more complex system using only \acp{BT}.


\section{High Level Modeling Language}\label{sec:modeling-language}

Robotic systems, especially the ones using \ac{ROS 2} and \acp{BT}, consist of a number of nodes running in parallel and exchanging information via communication interfaces, i.e., topics, services, and actions in \ac{ROS 2}, as well as ticks and blackboards in \acp{BT}.
In the following, we describe how these systems can be modelled.

We introduce \ac{HL-SCXML}, a custom modeling language building on top of \ac{SCXML}.
It supports \ac{ROS 2} communication interfaces and timers, and \ac{BT} functionalities such as input-output ports, the blackboard, and ticks.
Each \ac{HL-SCXML} file represents one state machine: this is used to represent complete \ac{ROS 2} nodes and \ac{BT} nodes.
Additionally, we support reading the definition of the \ac{BT} in the XML format of the \emph{BehaviorTree.CPP}\footnote{\url{https://github.com/BehaviorTree/BehaviorTree.CPP}} library, so users must only model the behavior of the leaf nodes explicitly.
This enables an accurate representation of the actual robotic control architecture.

The \ac{AS2FM} tool translates \ac{HL-SCXML} to standard \ac{SCXML}~\cite{jimbarnettStateChartXML2015}, replacing all references to \ac{ROS 2} and \ac{BT} features with \ac{SCXML} events.
We provide an overview of the translations from \ac{HL-SCXML} to \ac{SCXML} in the next paragraphs but, for brevity, we omit a detailed description in this paper and refer to our tool's documentation\footnote{\url{https://convince-project.github.io/AS2FM/scxml-jani-conversion.html}}.
Using standard \ac{SCXML} as intermediate format limits the amount of functionalities we need to consider when generating the formal model of the system.
Afterwards, AS2FM converts the standard \ac{SCXML} model into a network of \acp{MDP}, that is the formalism used by \jani, to be formally verified.  

\subsection{ROS 2 Interfaces}\label{subsec:ros-interfaces}

\ac{ROS 2} nodes are modeled using the following interface definitions:

\begin{itemize}
    \item \emph{ROS Topics} are implemented by translating them to \ac{SCXML} events.
    The data in events is read from the declared ROS message type.
    \item \emph{ROS Services} are implemented by translating the request and response to separate events,
    where the data is read from the declared ROS types.
    \item \emph{ROS Actions} are implemented by translating the goal, feedback, and result to separate events. The data is read from the declared ROS types.
    They can be modeled in a multi-threaded architecture, where the action server uses multiple threads to handle multiple goals concurrently. Each thread is a separate SCXML state machine.
    \item \emph{ROS Timers} are realized by the introduction of a state machine serving as a global clock.
    This distributes the time information to all state machines that may declare callbacks to be triggered at different rates.
\end{itemize}

For a full syntax reference, refer to our documentation.\footnote{\url{https://convince-project.github.io/AS2FM/howto.html}}

\subsection{Behavior Tree Runtime}\label{subsec:bt-runtime}

\acp{BT} can be modeled using the following concepts:

\begin{itemize}
    \item The \emph{Runtime} interprets the BT.xml\footnote{\url{https://www.behaviortree.dev/docs/learn-the-basics/xml\_format/}} file and executes the tree.
    It is ticked at a configurable rate.
    Also the behavior of ticking based on the tree return state can be configured.
    \item All \emph{BT Nodes}, including control nodes and leaf nodes, are represented as state machines, using \ac{HL-SCXML}.
    The tick signal is provided via events.
    In AS2FM, we provide an implementation of the standard BT nodes. They can be used as they are or as a starting point for custom ones.
    \item The \emph{Blackboard} is a key-value storage to exchange data between nodes at runtime.
    BT nodes can read and write data, which is exchanged via events from and to a BT management state machine.
\end{itemize}

\section{Model Translation}\label{sec:model-translation}

The full system model must be translated from a set of \ac{SCXML} state machines to the {\jani} format for model checking.

\subsection{SCXML System of State Machines}\label{ssec:scxml}

Consider a set $M = \{M_1, M_2, \ldots, M_k\}$ of $k$ state machines, where each $M_i$ is specified in an \ac{SCXML}-file.
Each state machine $M_i$ can communicate with the other ones using the events $E$.

\def\pentry{p_{\texttt{entry}}}
\def\pexit{p_{\texttt{exit}}}
\def\none{\text{0}}
\def\emptylist{[~]}

Each state machine $M_i$ is defined as a tuple $M_i = (D_i, C_i, P_i, S_i, s_{0i}, T_i)$, where:
\begin{itemize}
    \item $D_i$ is the set of data variables in $M_i$.
    \item $C_i$ is the set of conditions that are defined for $M_i$. Each condition $c \in C_i$ is a boolean expression over the data from the variables in $D_i$ and the data from the received events in $E$. Formally $c : (D_i \cup E) \to \{\false, \true\}$.
    \item $P_i$ defines executable content defined in $M_i$.
    Each executable content $p \in P_i$ can modify the data variables $D_i$ or send events $e \in E$.
    Each executable content $p \in P_i$ can be defined as a sequence of operations $p = [o_1, o_2, \dots]$.
    Each step $o_x$ is either a function $o_f$ modifying data objects or an event $o_e$ being sent.
    A function can modify data entries of this state machine based on the content of other data entries or data from a received event $o_f : (D_i \cup E) \times D_i$.
    An operation may also be to send an event $o_e \in E$ to other state machines.
    \item $S_i$ is the set of states in $M_i$.
    States may contain executable content that is executed \textit{onentry} and \textit{onexit}.
    A state $s \in S_i$ may have a tuple $s = (\pentry, \pexit)$, with $\pentry \in P_i \cup \{ \emptylist \}$ and $\pexit \in P_i \cup \{ \emptylist \}$.
    \item $s_{0i} \in S_i$ is the initial state of $M_i$.
    \item $t \in T_i = (S_i \times S_i \times (C_i \cup \{\true\}) \times (E \cup \{\none\}) \times (P_i \cup \{ \emptylist \}))$ is the set of transitions in $M_i$. Each transition $t$ is defined as a tuple $t = (s, s', c, e, p)$, where:
    \begin{itemize}
        \item $s, s' \in S_i$ are the source and target states of the transition.
        \item $c \in C_i \cup \{\true\}$ is the condition that must be satisfied for the transition to be taken. $c$ is set to $\text{\true}$ if there is no condition on the transition.
        \item $e \in E \cup \{\none\}$ is the event that triggers the transition, or $\none$ if no event triggers the transition. Note that events may send data to this state machine, which we consider as read-only variables.
        \item $p \in P_i \cup \{ \emptylist \}$ is the executable content associated with the transition.
    \end{itemize}
\end{itemize}

State machines $M_i$ and $M_j$ may exchange data and influence each other's execution by sending events.
Specifically, an event $e \in E$ emitted by $M_i$ may be received by $M_j$ where it is processed according to $M_j$'s transitions.
Data contained in the event is considered as an update to $D_j$.
This means, the synchronization between state machines is defined implicitly by the events exchanged between them.
\Cref{fig:state-machines} shows an example of two state machines that exchange events.

\subsection{JANI Network of \aclp{MDP}}\label{ssec:jani}

The {\jani}~\cite{DBLP:conf/tacas/BuddeDHHJT17} format encodes a complete system as a network of \acp{MDP}, including a list of synchronized edges that can only be executed simultaneously.
In the remainder of this section, we describe how to compile a {\jani} model from a system of \ac{SCXML} state machines.

\def\M{\mathcal{M}}
\def\D{\mathcal{D}}
\def\P{\mathcal{P}}
\def\p{\rho}
\def\G{\Phi}
\def\g{\varphi}
\def\S{\Sigma}
\def\s{\varsigma}
\def\T{\mathcal{T}}
\def\t{\tau}
\def\A{\mathcal{A}}
\def\a{\alpha}
\def\Y{\Gamma}
\def\y{\gamma}

Let $\M = \{\M_1, \M_2, \ldots, \M_n\}$ be the set of $n$ \acp{MDP} defined in the {\jani} model.
$\D_g$ is the set of global data variables shared by all \acp{MDP} in $\M$.
This global data is used to implement the exchange of data associated with events in the state machines in $M$.
Each \ac{MDP} $\M_i$ is defined as a tuple $\M_i = (\D_i, \G_i, \S_i, \s_{0i}, \A_i, \T_i)$, where:

\begin{itemize}
    \item $\D_i$ is the set of data variables in $\M_i$.
    \item $\G_i$ is the set of guards.
    Each guard $\g \in \G_i$ is a function of local and global variables evaluating to a binary value $\g : \D_i \cup \D_g \to \{\false, \true\}$.
    \item $\S_i$ is the set of states in $\M_i$.
    \item $\s_{0i} \in \S_i$ is the initial state of $\M_i$.
    \item $\A_i$ is the set of actions in $\M_i$.
    \item $\T_i = \S_i \times \S_i \times (\G_i \cup \{\true\}) \times \A_i \times (\P_i \cup \none)$ is the set of transitions in $\M_i$. Each transition $\t \in \T_i$ is defined as a tuple $\t = (\s, \s', \a, \g, \p)$, where:
    \begin{itemize}
        \item $\s, \s' \in \S_i$ are the source and target states of the transition.
        \item $\g \in (\G_i \cup \{\true\})$ is the guard that must be satisfied for the transition to be taken.
        It may be set to $\text{\true}$ if there is no guard on the transition.
        \item $\a \in \A_i$ is the action that triggers the transition.
        \item $\p \in (\P_i \cup \none)$ is the assignment associated with the transition.
        Where $\P_i$ defines assignments that modify the data variables in $\D_i$ and $\D_g$.
        Each assignment $\p \in \P_i$ is a function $\p_f : \D_i \times \D_g \to \D_i \times \D_g$.
        $\p$ is set to $\none$ if there is no assignment on the transition.
    \end{itemize}
\end{itemize}

The network of \acp{MDP} includes a set of synchronizations $\Y$.
Each synchronization $\y \in \Y$ is defined as a tuple $\y = (\a_1, \a_2, \ldots, \a_n)$, where $\a_i \in \A_i$ is an action in the \ac{MDP} $\M_i$.
One or more $\a_i$ may be $\none$ indicating that the \ac{MDP} $i$ does not have an action in this synchronization.
All actions $\a_1, \a_2, \ldots, \a_n$ must be executed simultaneously and all of their guards must be satisfied for the synchronization to be taken.
\Cref{fig:mdps} shows an example of two \acp{MDP} that synchronize using the actions $a_{e2s}$ and $a_{e2r}$.

\subsection{Translation Algorithm}\label{ssec:translation}

\def\eventsenders{\texttt{event\_senders}}
\def\eventreceivers{\texttt{event\_receivers}}
\def\uniqueid{\texttt{unique\_id}()}

\begin{algorithm}
\caption{Translation from a set of \ac{SCXML} state machines to a network of \acp{MDP}.}\label{alg:translation}
\begin{algorithmic}[1]
\Require$M, E, \uniqueid$\label{alg:ln:unique_id}
\State$\eventsenders \gets \varnothing$\label{alg:ln:init_eventsenders}
\State$\eventreceivers \gets \varnothing$\label{alg:ln:init_eventreceiver}
\State$\M \gets \varnothing$\label{alg:ln:init_m}
\State$\Y \gets \varnothing$\label{alg:ln:init_y}
\For{$M_i \in M$}
    \State$\S_i \gets S_i$\label{alg:ln:states}
    \State$\D_i \gets D_i$
    \State$\s_{0i} \gets s_{0i}$\label{alg:ln:initial_state}
    \State$\T_i \gets \varnothing$
    \For{$t \in T_i$}\label{alg:ln:transitions}
        \State$\s \gets s$\label{alg:ln:source_state}
        \State$\g \gets c$
        \State$\a \gets \uniqueid$
        \If{$e \neq \none$}
            \State$\eventreceivers[e] \gets \a$\label{alg:ln:event_receiver}
        \EndIf{}
        \For{$o \in p$}
            \State$\s_o \gets \uniqueid$\label{alg:ln:intermediate_state}
            \If{$o \in E$}
                \State$\eventsenders[o] \gets \a$\label{alg:ln:event_sender}
                \State$\p_o \gets \emptylist$
            \Else\label{alg:ln:is_executable_content}
                \State$\p_o \gets o$
            \EndIf{}
            \State$\t_o \gets (\s, \s_o, \a, \g, \p_o)$
            \State$\T_i \gets \T_i \cup \{\t_o\}$
            \State$\S_i \gets \S_i \cup \{\s_o\}$
            \State$\s = \s_o$
            \State$\a \gets \uniqueid$
            \State$\g = \true$
        \EndFor{}
        \State$\s' \gets s'$\label{alg:ln:original_goal}
        \State$\t \gets (\s, \s', \a, \g, \emptylist)$\label{alg:ln:tail_transition}
        \State$\T_i \gets \T_i \cup \{\t\}$
    \EndFor\label{alg:ln:transitions_end}
    \State$\M_i \gets (\D_i, \S_i, s_{0i}, \T_i)$
    \State$\M \gets \M \cup \{\M_i$\}\label{alg:ln:add_mdp}
\EndFor{}
\For{$e \in E$}
    \State$\s_{e0} \gets \uniqueid$ \label{alg:ln:init_event_states}
    \State$\s_{e1} \gets \uniqueid$
    \State$\S_e \gets \{s_{e0}, s_{e1}\}$
    \State$\a_{s} \gets \uniqueid$
    \State$\t_{s} \gets (\s_{e0}, \s_{e1}, \a_{s}, \true, \emptylist)$
    \State$\y_{s} \gets (\dots, \a_{s}, \dots, \eventsenders[e], \dots)$\label{alg:ln:event_sender_sync}
    \State$\a_{r} \gets \uniqueid$
    \State$\t_{r} \gets (\s_{e1}, \s_{e0}, \a_{r}, \true, \emptylist)$
    \State$\y_{r} \gets (\dots, \a_{r}, \dots, \eventreceivers[e], \dots)$\label{alg:ln:event_receiver_sync}
    \State$\Y \gets \Y \cup \{\y_{s}, \y_{r}\}$
    \State$\M \gets \M \cup \{(\varnothing, \S_e, \s_{e0}, \{\t_{s}, \t_{r}\})\}$\label{alg:ln:add_event_mdp}
\EndFor{}
\State\Return$\M$, $\Y$\label{alg:ln:return}
\end{algorithmic}
\end{algorithm}

The translation is defined in \Cref{alg:translation}.
It requires a set of \ac{SCXML} state machines $M$ and the set of events $E$ that are exchanged between the state machines.
It further requires a function $\uniqueid$ that returns a unique identifier for each element in the model in line~\ref{alg:ln:unique_id}.
The data structures $\eventsenders$ and $\eventreceivers$ are used to keep track of the events that are sent and received by the state machines and are initialized empty in lines~\ref{alg:ln:init_eventsenders} and~\ref{alg:ln:init_eventreceiver}.
Also the set of \acp{MDP} $\M$ and the set of synchronizations $\Y$ in lines~\ref{alg:ln:init_m} and~\ref{alg:ln:init_y} are initialized empty.

Then, per state machine $M_i$ in $M$, the states $S_i$, data variables $D_i$, and initial state $s_{0i}$, are copied to the \ac{MDP} $\M_i$ (lines~\ref{alg:ln:states} to~\ref{alg:ln:initial_state}).
The translation of transitions and their executable content $p$ happens in lines~\ref{alg:ln:transitions} to~\ref{alg:ln:transitions_end}.
Unique actions get generated and in case events are to be received or sent, they are stored (lines~\ref{alg:ln:event_receiver} and~\ref{alg:ln:event_sender}).
The \ac{MDP} $\M_i$ gets an additional intermediate state $\s_o$ per operation $o$ of the edge's executable content $p$ (line~\ref{alg:ln:intermediate_state}).
The same is done for executable content in $\pentry$ and $\pexit$, detailed description is omitted for brevity.
After all transitions are generated, the \ac{MDP} $\M_i$ is added to the set of \acp{MDP} $\M$ in line~\ref{alg:ln:add_mdp}.

For each event $e$ in $E$, we create a new \ac{MDP} consisting of two states $\s_{e0}$ and $\s_{e1}$ (lines~\ref{alg:ln:init_event_states} - \ref{alg:ln:add_event_mdp}).
$\s_{e1}$ is the active state when an event $e$ is present and needs to be processed on the receiving side, $\s_{e0}$ is active otherwise.
Therefore, we add a transition from $\s_{e0}$ to $\s_{e1}$ that is synchronized with the action that sends the event $e$ in line~\ref{alg:ln:event_sender_sync}.
Similarly, we add a transition from $\s_{e1}$ to $\s_{e0}$, synchronized with the action receiving it in line~\ref{alg:ln:event_receiver_sync}.
Finally, the additional \ac{MDP} $(\varnothing, \S_e, \s_{e0}, \{\t_{s}, \t_{r}\})$ is added to the set of \acp{MDP} $\M$ (line~\ref{alg:ln:add_event_mdp}) before returning $\M$ and $\Y$ (line~\ref{alg:ln:return}).

\section{Evaluation}\label{sec:evaluation}

In this section we evaluate the most important aspects of our approach.
Firstly, we show a case study by verifying the correctness of a full robotic system after generating its formal model with AS2FM.
Then, we evaluate the model checking runtime on that system across different SMC confidence intervals.
Finally, we compare our toolchain consisting of AS2FM and a statistical model checker against the state-of-the art, BehaVerify~\cite{serbinowskaFormalizingStatefulBehavior2024}, in terms of runtime scalability.

We use the \ac{SMC}-tool \emph{SMC Storm}~\cite{lampacresciaVerifyingRoboticSystems2024} with a confidence of $95\%$ and a maximum error of $0.01$, unless specified differently.
All \ac{SMC}-tools use these two values to define the desired confidence interval, which in turn determines the required amount of traces to generate during the verification process.
All experiments are performed with an i7-11850H CPU and 32 GB of memory.

\subsection{Case Study}\label{sec:case-study}

\begin{figure}
    \centering
    \includegraphics[width=.65\columnwidth]{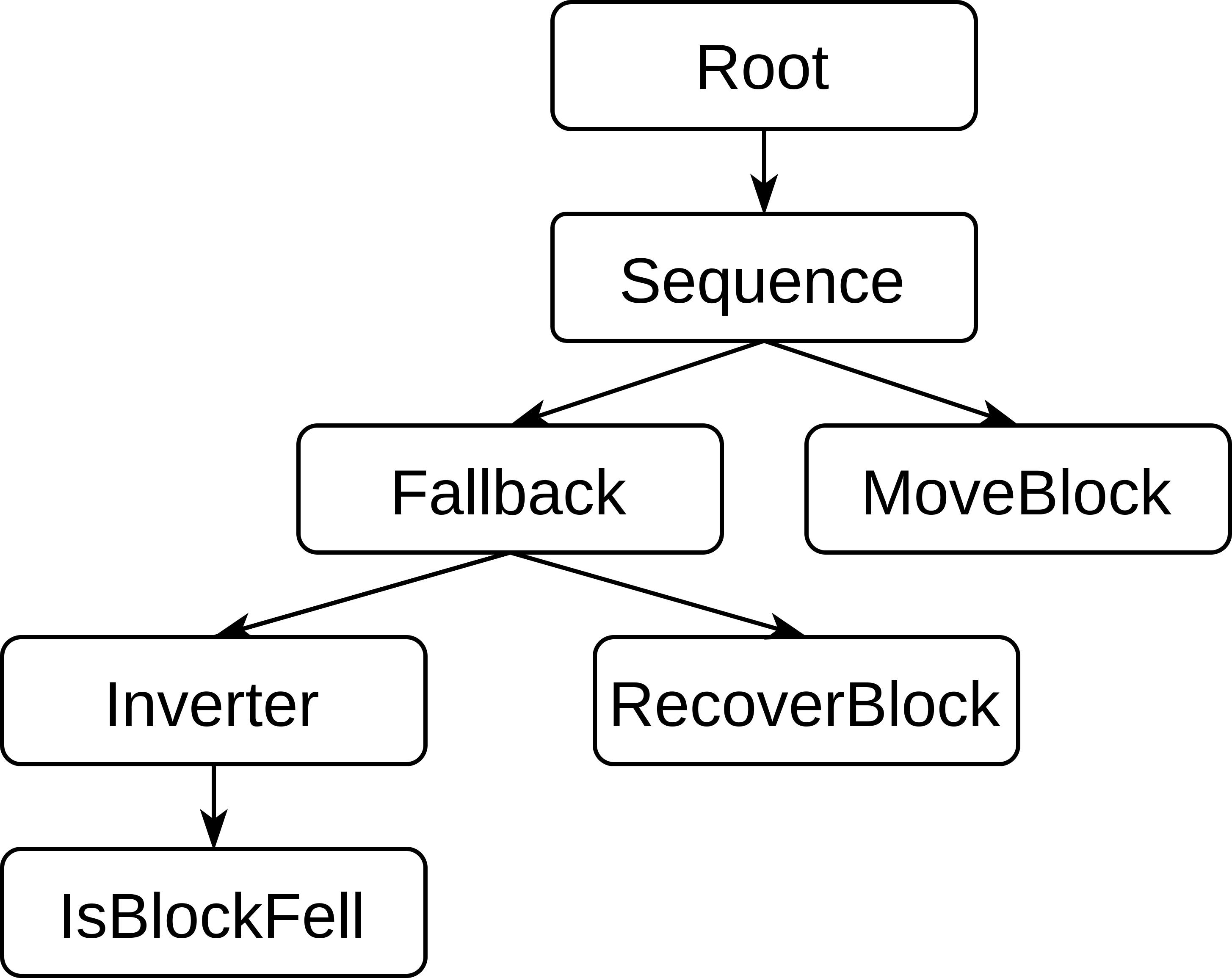}
    \caption{The \ac{BT} used for controlling the assembly robot in the case study.}
    \label{fig:case-study-bt}
    \vspace*{-0.2cm}
\end{figure}

This section demonstrates how our tooling is applied to a robotic control system based on \acp{BT} and \ac{ROS 2}.
The goal is to verify that an autonomous assembly robot reacts correctly in case a block falls from its gripper.
The model consists of $14$ separate SCXML state machines, including:
\begin{itemize}
    \item The \ac{BT} executing the policy shown in \Cref{fig:case-study-bt}.
    \item A model of the user-defined \ac{BT} plugins: IsBlockFell, MoveBlock, RecoverBlock.
    \item \ac{ROS 2} nodes encoding specific robot skills triggered from the \ac{BT}: MoveBlockSkill, RecoverBlockSkill.
\end{itemize}

The system is expected to tick the \ac{BT} root for as long as it returns \emph{RUNNING}.
Our goal is to verify that the move block operation is successful or, in case the block operation is aborted, the block recovery skill is enabled in time.
This is described in \ac{LTL}~\cite{DBLP:conf/banff/Vardi95} as:
\def\abort{\texttt{abort}}
\def\success{\texttt{success}}
\def\recovery{\texttt{recovery}}
\begin{align}
    &(\abort \implies (t_{curr} < (t_{abort} + t_{timeout})))~ \nonumber \\
    &U (\success \lor \recovery),
    \label{eq:uc2-ltl}
\end{align}
where $\abort$ and $\success$ are boolean flags for representing if the move action was aborted or successful, $\recovery$ is a flag signaling if the recovery sequence was started, $t_{curr}$ is the current time, $t_{abort}$ is the time the move action was aborted, and $t_{timeout}$ is a constant representing the maximum time the system can take to start the recovery sequence.
$U$ is an LTL operator that reads as \emph{until} and is true if the left side is true in at least all states before the state in which the right side became true for the first time.

After converting the model of this robot to {\jani} with AS2FM, and giving it to \ac{SMC} Storm, the model checker reports that the probability of the model satisfying the property from \Cref{eq:uc2-ltl} is $0.499$: this means that there are cases in which the system is not behaving as expected.

\ac{SMC} tools offer the possibility of exporting the generated traces as a CSV file, for inspecting them and determining the reason for the failure.
In this particular case, the \ac{BT} checks if the block fell only before starting to move it. It is unable to detect this after the MoveBlock plugin is started.
This results in the system being unable to start the recovery sequence in case the block falls from the gripper, which in the model we used happens $50\%$ of the times.

The issue can be solved by replacing the \emph{Sequence} control node with a \emph{ReactiveSequence} one, that keeps checking if the blocks fell and, if that is the case, starts a recovery.
Once the \ac{BT} is fixed, the model can be checked again: this time the tool reports that the probability of satisfying the property is $1$, confirming that the implementation is working as expected.

\subsection{Runtime over Confidence Value}\label{sec:confidence-intervals}
Again on the previous case study, we evaluate how different values for the confidence affect the number of necessary traces and therefore the runtime of the \ac{SMC} tool.
For each confidence value, we run the verification $100$ times and measure the average runtime.

\def\width{.8\columnwidth}
\def\trimtop{40pt}

\begin{figure}[t]
    \centering
    \includegraphics[width=\width,trim={0 0 0 \trimtop},clip]{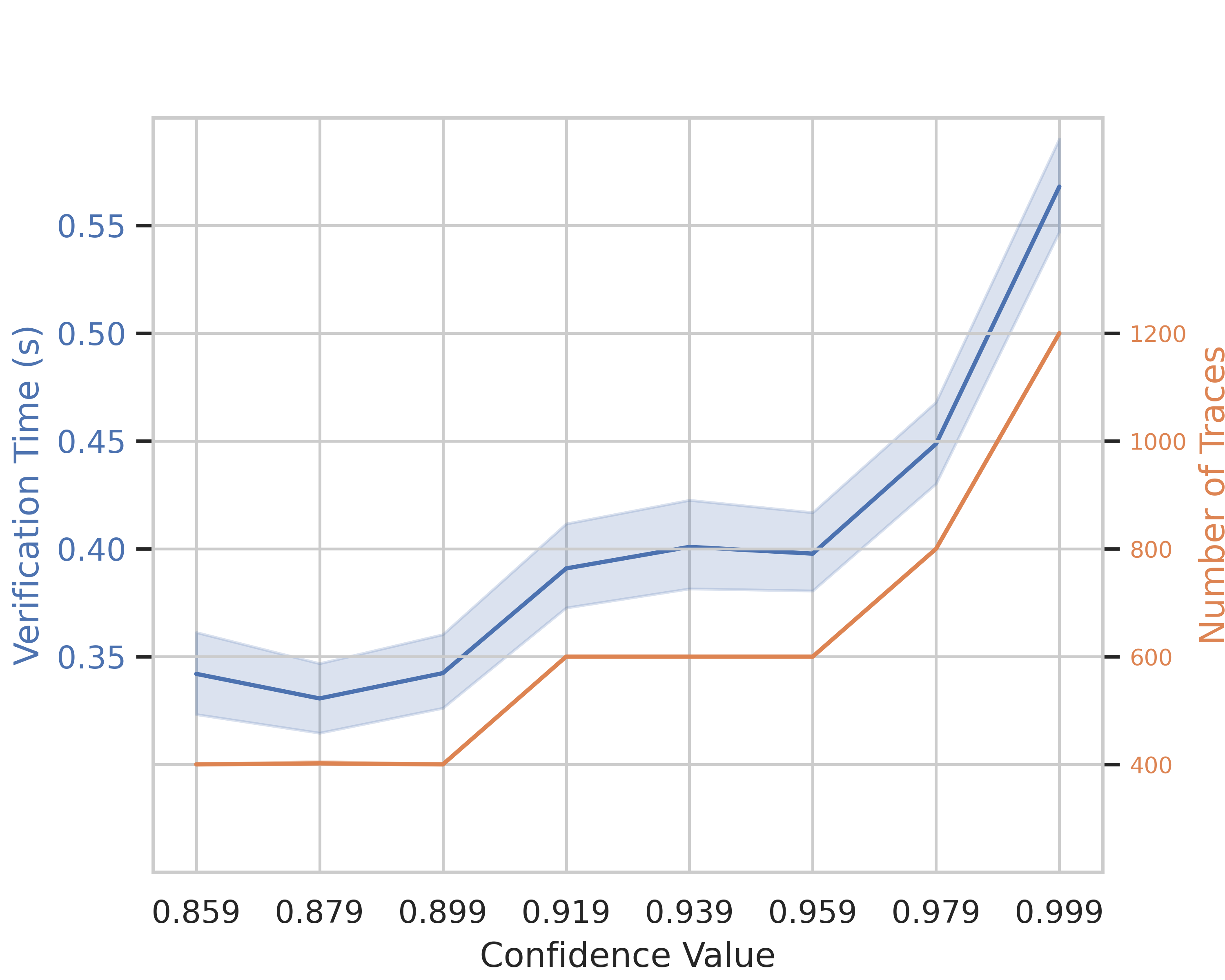}
    \caption{Comparing runtime and number of traces for different confidence intervals. Both metrics increase with the confidence interval.}\label{fig:performance-confidences}
\end{figure}
\begin{figure}[t]
    \centering
    \includegraphics[width=\width,trim={0 0 0 \trimtop},clip]{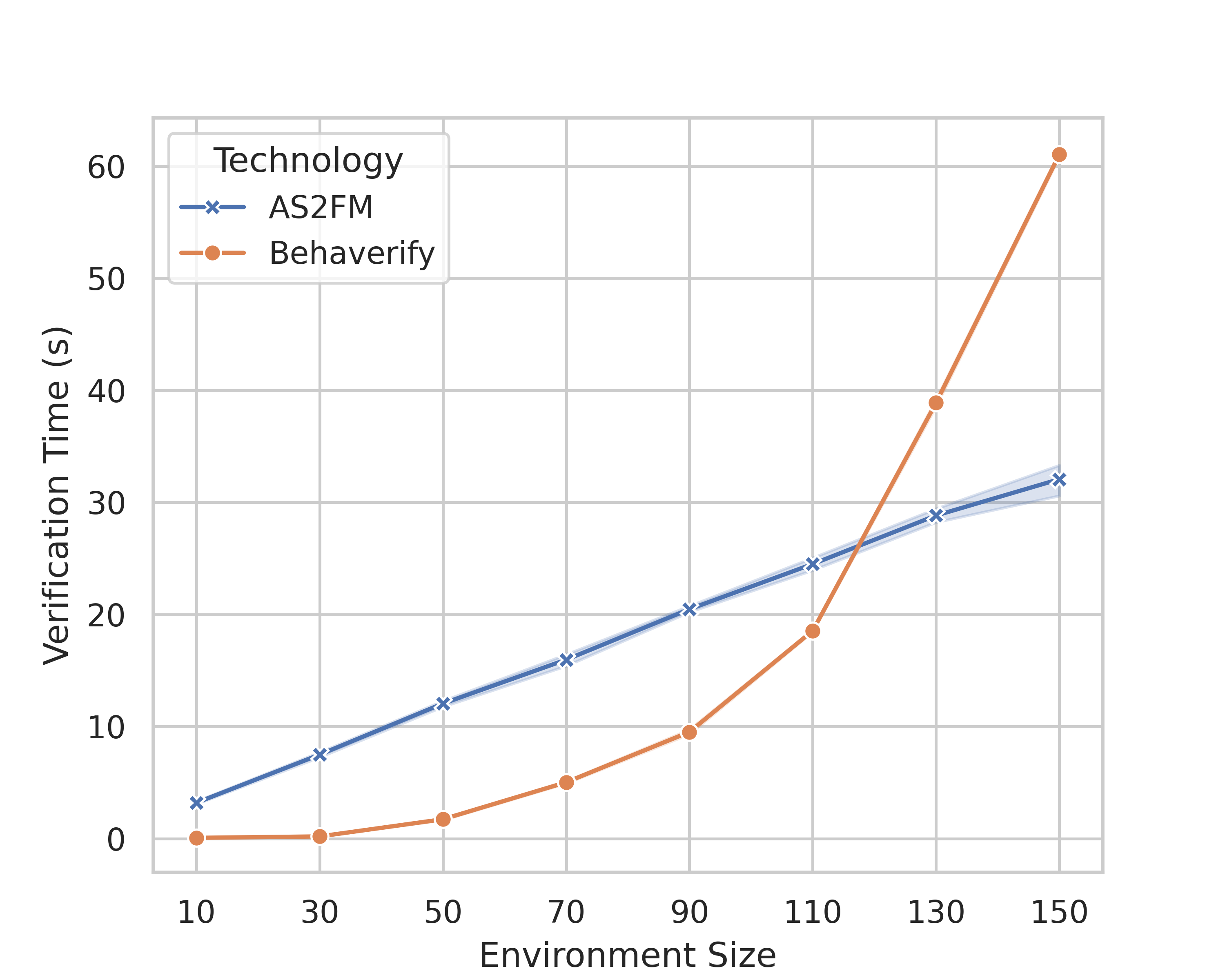}
    \caption{Performance and scalability of the AS2FM + SMC Storm toolchain compared to BehaVerify. It is visible that the runtime of our tooling scales linearly with the size of the grid, while BehaVerify scales exponentially.}\label{fig:performance-scalability}
    \vspace*{-0.2cm}
\end{figure}

It is visible from the results in \Cref{fig:performance-confidences} that the runtime of SMC Storm increases with the confidence interval.
This is expected, as the tool needs to generate more traces to reach the desired confidence.
The number of traces generated also increases with the confidence interval, which is the reason for the increased runtime.

\subsection{Runtime and Scalability}\label{sec:performance-scalability}

The runtime scalability of the AS2FM toolchain was evaluated in direct comparison to \emph{BehaVerify}~\cite{serbinowskaFormalizingStatefulBehavior2024}, the state-of-the-art framework for verifying \acp{BT} only, which does not provide robotic-specific features, e.g., ROS communication, but which is the closest tool to ours.
We used the example in their paper called \emph{Simple Robot}, based on a 4-connected grid world with a robot that has to reach randomly placed goal locations.
The robot is controlled by a \ac{BT}.
To scale this example, the size of the grid in both dimensions is changed to the indicated \texttt{size}.
This example was chosen because the other scaling example in their work, \emph{Bigger Fish}, is perceived as being specifically designed to work well with their custom optimization and does therefore not provide insights into the general performance of the verification tooling.

In \Cref{fig:performance-scalability} it can be seen, that the runtime of our tool is larger than that of BehaVerify for sizes of 110 or less.
This can be attributed to the overhead that is introduced in our much more general approach by a more sophisticated model of communication (e.g., for ROS) which is crucial for the applicability in real-world robotic systems.
But it is further visible that the runtime of our approach scales linearly with the size of the environment, while BehaVerify scales exponentially.
These results can be attributed to using \ac{SMC}  opposed to full-state space model checking, which further underlines the practical usability of our approach with real-world systems that grow quickly in size.

\section{Conclusion}\label{sec:conclusion}

We have demonstrated our approach for using \acl{SMC} on full models of robotic control systems to increase their robustness.
The models consist of SCXML state machines that benefit from an extension allowing the direct modelling of ROS 2 communication interfaces, and \aclp{BT} in their native XML syntax.
We contribute an algorithm to translate the full model in a network of \acp{MDP}.
We demonstrated the approach on an industrial case study successfully identifying errors, and showing better scalability compared to the state of the art.
In the future, the approach can be extended and improved by adding support for more complex environment models, by providing tools for easier inspection of the generated SMC traces and model checking results, and by making the SCXML format more accessible with the help of graphical interfaces.


\bibliographystyle{IEEEtran}
\bibliography{references,refs-non-zotero}

\end{document}